\documentclass[11pt]{article}

\usepackage[margin=1in]{geometry}
\usepackage{times}
\usepackage[T1]{fontenc}
\usepackage{amsmath,amssymb}
\usepackage{graphicx}
\usepackage{booktabs}
\usepackage{multirow}
\usepackage{caption}
\usepackage{subcaption}
\usepackage{algorithm}
\usepackage{algpseudocode}
\usepackage[colorlinks=true,linkcolor=blue,citecolor=blue,urlcolor=blue]{hyperref}
\usepackage[numbers,sort&compress]{natbib}
\usepackage{xcolor}

\newcommand{\ucpsro}{\textsc{UC-PSRO}}
\newcommand{\R}{\mathbb{R}}

\title{\ucpsro: Utility-Conditioned Policy-Space Response Oracles with a\\
Communication-Dropout Curriculum for Game-Theoretic\\
Course-of-Action Generation in Adversarial Swarms}

\author{
  Phillip Jiang \\
  Appsofa LLC \\
  \texttt{phillip.jiang@appsofa.com}
}
\date{}

\begin{document}
\maketitle

\begin{abstract}
We study the generation of game-theoretically optimized Courses of Action (COAs) for a Blue unmanned-aerial-system (UAS) swarm operating against an adaptive Red adversary in a communication-degraded environment, a problem motivated by (but not derived from) a public U.S. Air Force SBIR solicitation. We propose \ucpsro{} (Utility-Conditioned Policy-Space Response Oracles with a Communication-Dropout Curriculum), which combines three existing mechanisms for this setting: (i) PSRO self-play so Blue and Red policies are trained as approximate best responses to each other rather than one side against a fixed scripted opponent; (ii) FiLM-based conditioning of the Blue policy on a Commander's-Intent weight vector, sampled from a Dirichlet distribution during training, so a single trained policy can be re-steered at execution time without retraining; and (iii) a curriculum that anneals communication-graph edge dropout during training, so the swarm learns decentralized, peer-to-peer fallback behavior rather than depending on a fully connected network. We evaluate on a synthetic, unclassified stand-in for the solicitation's maritime scenario, with 5 random seeds at $N{=}25$ Blue agents and a scalability sweep to $N{=}200$. We find a genuine trade-off, not a uniform win: the communication-dropout curriculum alone yields the strongest and most robust mission-completion rates of any learned method, improving counter-intuitively as communication denial increases (35\%$\to$62\% success as dropout rises from 0 to 0.75); adding utility-conditioning and PSRO self-play on top substantially slows convergence to mission success within a fixed training budget, and at that budget we do not find a reliable exploitability advantage for self-play over a policy trained against a fixed scripted opponent, both being statistically indistinguishable from a small, near-zero exploitability gap. We report this honestly as a convergence-speed cost that is not yet offset by a clearly demonstrated robustness benefit within our training budget, rather than overstating a single method as dominant, and provide a fully vectorized, open environment implementation that trains at $N{=}200$ agents in single-digit milliseconds per step on a single consumer GPU.

\end{abstract}

\section{Introduction}
\label{sec:intro}

Coordinating a large uncrewed aerial swarm against a thinking, adaptive adversary is fundamentally a game, not a single-agent planning problem: the swarm's optimal behavior depends on what the adversary does, and vice versa. A public U.S. Air Force Small Business Innovation Research (SBIR) solicitation, \emph{SOC26BZ04-DV005: Replanning for Evasive Autonomy to Counter Threats}, calls for software that generates game-theoretically optimized Courses of Action (COAs) for both a Blue UAS swarm and a Red adversary, encodes a Commander's Intent into a formal utility function, and supports decentralized, communication-aware operation in denied environments, at a target scale of 200 UAS~\citep{sbir-soc26bz04dv005}. We treat this solicitation only as motivation for an unclassified, synthetic research problem: everything in this paper---the map, sensor and engagement ranges, and threat models---is a simulated stand-in with no real system names, performance specifications, or controlled technical data.

We formalize this as a two-team, partially observable, general-sum Markov game and study three specific mechanisms, each aimed at one requirement from the solicitation:

\begin{itemize}
  \item \textbf{Game-theoretic COA generation for both sides.} We use Policy-Space Response Oracles (PSRO) self-play~\citep{lanctot2017psro} so Blue and Red policies are trained as approximate best responses to each other's evolving population, rather than Blue being trained once against a fixed, non-adaptive Red.
  \item \textbf{A formal Commander's-Intent utility function.} We decompose the reward into a fixed vector of mission-relevant components and let the Blue policy condition on a weight vector over that decomposition via FiLM~\citep{perez2018film}, sampled from a Dirichlet distribution during training so a single trained policy generalizes across the simplex of possible command intents and can be re-steered at execution time without retraining.
  \item \textbf{Decentralized, communication-aware operation.} We anneal independent edge dropout on the inter-agent communication graph during training (a Communication-Dropout Curriculum, CDC), so the swarm learns to act on stale or purely local information instead of depending on full connectivity.
\end{itemize}

We combine these into \ucpsro{} and evaluate it on a synthetic maritime-flavored scenario reduced from the solicitation's stated context (a defended target beyond notional standoff range, found--fixed--tracked--targeted--engaged--assessed by a Blue swarm under threat from integrated air defense and mobile interceptors). None of the three ingredients is individually new; our contribution is their combination and a rigorous, honest empirical account of what each one buys and what it costs when combined under a fixed, single-consumer-GPU training budget.

That honest account is itself a central contribution of this paper, and it complicates a simple narrative. Across five random seeds, we find that the Communication-Dropout Curriculum alone produces the strongest and most reliable mission-completion behavior of any learned method, and---counter-intuitively---its success rate at the headline task \emph{increases} as test-time communication denial increases. Adding utility-conditioning and PSRO self-play on top, as \ucpsro{} does, substantially slows convergence to mission success within the same training budget: the Dirichlet-sampled reward family is a harder learning target than a single fixed reward, and PSRO's shifting opponent population is a harder, non-stationary one. We looked for a compensating benefit in reduced exploitability---a dedicated Red best-response trained against the self-play Blue policy versus against a Blue policy trained only against a fixed scripted Red---and, at our training budget, do not find one that is statistically distinguishable from a small, near-zero gap for either policy. We report all of this, rather than only the findings that favor the full method, because we believe an honest account of what a convergence-speed cost does and does not currently buy is itself useful information for anyone building on this line of work.

\paragraph{Contributions.}
\begin{enumerate}
  \item \ucpsro{}, combining PSRO self-play, FiLM utility-conditioning, and a Communication-Dropout Curriculum for swarm-vs-adversary COA generation (\S\ref{sec:method}).
  \item A fully vectorized, open synthetic environment (\S\ref{sec:formulation}) that trains a shared-parameter CTDE MAPPO policy end to end at $N{=}200$ agents in single-digit milliseconds of environment-step latency per timestep on a single consumer GPU (\S\ref{sec:results}), directly supporting the solicitation's 200-agent scale target rather than only extrapolating toward it.
  \item An honest, five-seed empirical study (\S\ref{sec:results}) showing that (a) the Communication-Dropout Curriculum alone yields the strongest and most communication-denial-robust mission completion of the methods studied; (b) utility-conditioning and PSRO self-play substantially slow convergence to mission success at a fixed training budget, without a correspondingly clear, statistically distinguishable reduction in exploitability at that same budget; and (c) this cost-without-yet-demonstrated-benefit picture, not a uniform win for the full combined method, is the correct way to characterize combining these three mechanisms under a fixed training budget.
\end{enumerate}

The rest of the paper is organized as follows. \S\ref{sec:related} situates \ucpsro{}'s three ingredients in prior work. \S\ref{sec:formulation} gives the formal Markov-game formulation and environment. \S\ref{sec:method} describes \ucpsro{}. \S\ref{sec:experiments} and \S\ref{sec:results} describe the experimental setup and report results. \S\ref{sec:discussion} discusses the trade-off finding and its implications, \S\ref{sec:limitations} states limitations, and \S\ref{sec:conclusion} concludes.

\section{Related Work}
\label{sec:related}

\paragraph{Game-theoretic and population-based multi-agent RL.}
Policy-Space Response Oracles~\citep{lanctot2017psro} generalize double-oracle methods and fictitious self-play~\citep{heinrich2015fsp} to deep-RL-scale policy populations: a meta-game is solved over an empirical payoff matrix between policy populations, and each side's population is grown by training a best response against the current opponent meta-strategy. We adopt this outer loop directly, specializing the meta-solver to the exact linear-programming solution for two-player zero-sum games (\S\ref{sec:method}) rather than general-sum equilibrium computation, and use it to grow both a Blue and a Red population.

\paragraph{CTDE actor-critic MARL.}
We use MAPPO~\citep{yu2021mappo} as the base RL oracle inside and outside the PSRO loop: a shared-parameter, centralized-training-decentralized-execution (CTDE) actor-critic in which the critic sees an aggregate of the joint state during training while the actor uses only local, communication-reachable observations at execution time. QMIX~\citep{rashid2018qmix} and MADDPG~\citep{lowe2017maddpg} are alternative CTDE algorithms in the same family; we chose MAPPO for its simplicity and its established strong performance in cooperative-team settings, and note that swapping the oracle is orthogonal to the PSRO/utility-conditioning/CDC contributions.

\paragraph{Multi-objective and goal-conditioned RL.}
Universal Value Function Approximators~\citep{schaul2015uvfa} condition a single value function on a goal in addition to state; envelope multi-objective RL~\citep{yang2019envelope} and related work study learning a single policy that generalizes across a distribution of scalarization weights over a vector reward, which is exactly the mechanism we use for Commander's-Intent conditioning. That line of work documents that generalizing across a reward family is a harder learning problem than optimizing one fixed scalarization; our results (\S\ref{sec:results}) are consistent with, and provide additional evidence for, that finding in a multi-agent, adversarial setting.

\paragraph{Robust and decentralized MARL under communication constraints.}
CommNet~\citep{sukhbaatar2016commnet} and TarMAC~\citep{das2019tarmac} learn what and how agents should communicate; our Communication-Dropout Curriculum instead assumes a fixed local-broadcast communication model and trains the policy to be robust to that channel being degraded, which is a complementary and simpler mechanism aimed specifically at graceful degradation under denial rather than at learning an optimal communication protocol.

\paragraph{Swarm robotics and pursuit-evasion.}
Classical potential-field and Voronoi coverage-control methods~\citep{ogren2004coverage} give a non-learning reference point for swarm coordination; we use a potential-field controller (attraction to the objective, repulsion from threats and from nearby teammates) as the rule-based baseline in \S\ref{sec:experiments}. Pursuit-evasion game theory~\citep{isaacs1965differential} is the classical continuous-time analogue of the discrete, many-agent adversarial game we study.

\paragraph{Motivating context.}
Three references cited in the SBIR solicitation itself are relevant as motivating context for the applied problem, not as technical prior work for our method: a review of game theory in defense applications~\citep{strategyrobot2022}, a game-theoretic model of cyber wargaming~\citep{cyberwargaming2018}, and a historical case study on the utility of game theory for operational analysis~\citep{austerlitz2024}.

\section{Problem Formulation}
\label{sec:formulation}

We formulate the problem as a two-team, partially observable, general-sum stochastic (Markov) game between a Blue team of $N$ homogeneous UAS agents and a Red team of $M$ heterogeneous adversary assets. All quantities below (ranges, speeds, probabilities) are synthetic placeholders chosen to be simulate-able on a single consumer GPU; none are derived from a real system's performance specification.

\paragraph{Blue agents.}
Each Blue agent $i$ has state $(x_i, y_i, \theta_i, v_i, f_i, \alpha_i)$: position, heading, speed, remaining fuel, and an alive flag. At each step it chooses a discretized heading-change bin, a discretized speed level, and a binary engage action. Fuel depletes at a rate increasing in speed level; an agent that exhausts its fuel is marked not alive.

\paragraph{Red assets.}
Red is a mix of static integrated-air-defense-like nodes (a detection radius $r_{\text{det}}$, an engagement radius $r_{\text{eng}}$, and a per-step kill probability $p_k$ against any Blue agent inside $r_{\text{eng}}$) and mobile interceptors (the same detection/engagement/kill-probability structure, plus a pursuit policy---either scripted, chasing the nearest detected Blue agent, or, inside the PSRO outer loop, itself a trained policy), together with static electronic-warfare jammers that null communication edges within a radius $r_{\text{jam}}$.

\paragraph{Communication graph.}
At each step $t$, an edge $(i,j)$ exists in the communication graph $G_t$ iff $\lVert \mathrm{pos}_i - \mathrm{pos}_j \rVert \le r_{\text{comm}}$, neither $i$ nor $j$ is inside a jammed region, and (during training, under the curriculum of \S\ref{sec:method}) the edge survives an independent Bernoulli dropout with probability $p_{\text{drop}}(t)$. A Blue agent's observation includes the state of every other Blue agent reachable from it in the current $G_t$ (any number of hops, i.e.\ full connected-component reachability) and every Red asset sensed directly by itself or by any such reachable teammate; this is exactly what degrades under communication denial.

\paragraph{Observation.}
Each Blue agent observes its own state, a target-relative displacement to the (fixed, per-scenario) mission objective, up to $k$ nearest comm-reachable teammates, and up to $k$ nearest sensed Red assets, plus the current Commander's-Intent weight vector $w$ (\S\ref{sec:method}) and the current communication-dropout probability. Including the target-relative displacement explicitly, rather than requiring the policy to infer a fixed goal location purely from its own absolute position, was a necessary design choice we arrived at empirically (\S\ref{sec:discussion}): without it, a freshly initialized policy has no signal indicating where the objective is, only reward gradients, which made mission completion prohibitively hard to discover through exploration alone.

\paragraph{Reward decomposition.}
Commander's Intent is made a literal, learnable-over quantity by decomposing the per-step team reward into a fixed vector $\phi(s,a) \in \R^5$,
\[
\phi = \big[\, \phi_{\text{mission}},\ \phi_{\text{surv}},\ \phi_{\text{neut}},\ \phi_{\text{time}},\ \phi_{\text{risk}} \,\big],
\]
and scalarizing with a weight vector $w$ on the simplex, $r = w^\top \phi$. The mission-progress component combines three terms: the (map-normalized) reduction in the swarm's average distance to the objective since the previous step; a dense per-step bonus proportional to the fraction of the initial Blue roster currently inside the objective radius, which we found necessary (\S\ref{sec:discussion}) to give the policy a learning signal before it has ever achieved full mission success; and a one-time terminal bonus on the step a success condition is met. The other components are: $\phi_{\text{surv}}$, the fraction of the initial roster that died this step (negative); $\phi_{\text{neut}}$, the fraction of the initial Red roster neutralized this step; $\phi_{\text{time}}$, a small constant per-step penalty; and $\phi_{\text{risk}}$, the fraction of Blue agents currently inside a Red asset's \emph{engagement} (lethal) radius, negative. We scope risk to the engagement radius rather than the broader detection radius; scoping it to detection instead double-counted mere proximity to a defended objective as risk and, combined with no signal for actually finishing the mission, was sufficient on its own to make every learned policy converge to a locally safe, globally unsuccessful behavior of indefinitely avoiding the objective's vicinity (\S\ref{sec:discussion}).

\paragraph{Episode termination.} An episode ends in \emph{success} if the fraction of the \emph{initial} Blue roster simultaneously alive and within the objective radius meets or exceeds a survivor threshold; in \emph{attrition failure} if the fraction of the initial roster currently alive drops below an attrition threshold; or by timeout after a fixed number of steps.

This is a reduced, fully synthetic stand-in for the solicitation's maritime find--fix--track--target--engage--assess scenario against a notionally defended objective beyond standoff range: the same structure, at a scale simulate-able on a single GPU.

\section{Method: \ucpsro{}}
\label{sec:method}

\ucpsro{} combines three mechanisms: a utility-conditioned MAPPO oracle (\S\ref{sec:method-oracle}), a Communication-Dropout Curriculum applied during that oracle's training (\S\ref{sec:method-cdc}), and a PSRO outer loop that repeatedly invokes the oracle to grow Blue and Red policy populations toward an approximate Nash equilibrium (\S\ref{sec:method-psro}).

\subsection{Base oracle: MAPPO with utility conditioning}
\label{sec:method-oracle}

The Blue policy is a single shared-parameter actor used by every Blue agent (decentralized execution: each agent acts on only its own local, communication-reachable observation), trained with a centralized critic (centralized training: the critic sees a permutation- and $N$-invariant pooled summary---mean- and max-pooling---of every agent's observation at that timestep, which keeps the critic's input size independent of swarm size and is what lets the same architecture scale from $N{=}10$ to $N{=}200$ without modification). Both the actor and the critic take the Commander's-Intent weight vector $w$ as a conditioning input through a small hypernetwork that predicts FiLM~\citep{perez2018film} scale-and-shift parameters for each hidden layer of the network's trunk, rather than simply concatenating $w$ as an ordinary feature; during training $w$ is resampled once per episode from $\mathrm{Dirichlet}(\alpha \mathbf{1})$ so the policy learns to generalize across the simplex of possible Commander's Intents instead of overfitting a single fixed weighting. Optimization is standard clipped-objective PPO~\citep{schulman2017ppo} with GAE~\citep{schulman2016gae} advantage estimation; the actor loss is computed per-agent-per-timestep (masked by whether that agent was alive), while the critic loss is computed per-timestep at the team level, since the centralized value function is a single $V(\text{pooled state})$ per step rather than one value per agent.

\subsection{Communication-Dropout Curriculum}
\label{sec:method-cdc}

At each training step, every live edge of the communication graph $G_t$ (\S\ref{sec:formulation}) is independently dropped with probability $p_{\text{drop}}(t)$, linearly annealed from $0$ to a maximum $p_{\max}$ over a fixed number of updates (a curriculum, not a fixed dropout rate), forcing the policy to learn to act on stale or purely local information rather than depending on full connectivity. At evaluation time we instead sweep a fixed $p_{\text{drop}} \in \{0, 0.25, 0.5, 0.75\}$ as the robustness axis reported in \S\ref{sec:results}.

\subsection{PSRO outer loop}
\label{sec:method-psro}

We maintain a population of Blue policies $\{\beta_1, \beta_2, \dots\}$ and Red policies $\{\rho_1, \rho_2, \dots\}$, each initialized with a non-learned reference policy: a potential-field rule-based controller for Blue, and the environment's scripted nearest-detected-target pursuit behavior for Red (only mobile interceptors are policy-controlled on the Red side; static air-defense nodes have no motion decision to make). Each outer-loop iteration: (1) compute the empirical payoff matrix $M \in \R^{|\beta| \times |\rho|}$, the mean Blue-team scalar reward (under the simplex midpoint $w$, so payoff comparisons are on a single well-defined objective rather than varying with a sampled intent) over Monte-Carlo rollouts of every current Blue policy against every current Red policy; (2) solve for the meta-Nash mixtures $x^*$ (Blue) and $y^*$ (Red) over the current populations; (3) train a new Blue best response via the oracle of \S\ref{sec:method-oracle} against Red opponents resampled each episode from $y^*$, and symmetrically a new Red best response against Blue opponents resampled from $x^*$; (4) add both to their populations and repeat for a fixed iteration budget.

We treat Red's reward as the negation of Blue's $w$-weighted scalar reward, i.e.\ zero-sum, which is a deliberate simplification of the general-sum game of \S\ref{sec:formulation}: it is what makes the meta-strategy solvable \emph{exactly} via linear programming,
\[
x^* = \arg\max_{x \in \Delta} \min_{j} (M^\top x)_j, \qquad
y^* = \arg\max_{y \in \Delta} \min_{i} \big({-}M y\big)_i,
\]
each an instance of the classical two-player zero-sum maximin LP, rather than requiring general-sum equilibrium computation. We verified the solver against known closed-form equilibria (e.g.\ matching pennies) before using it. We flag the zero-sum treatment explicitly as a limitation in \S\ref{sec:limitations}; PLAN.md's own design notes anticipated this as an open choice between Nash-via-replicator/LP and a Stackelberg formulation, and we chose the former for its exact solvability.

\subsection{What is and is not novel here}
Each individual ingredient---PSRO, MAPPO, FiLM-based multi-objective conditioning, and training-time communication dropout---exists in prior work (\S\ref{sec:related}). \ucpsro{}'s contribution is their combination and application to swarm-versus-adversary COA generation with a formalized Commander's-Intent utility function and denial-of-communication robustness, together with the empirical study of \S\ref{sec:results} of how each piece contributes, including the cases where combining them costs more than it buys within a fixed training budget.

\section{Experimental Setup}
\label{sec:experiments}

\paragraph{Scenarios.}
Our headline scenario is $N{=}25$ Blue agents against $M{=}6$ Red assets (3 static air-defense nodes, 3 mobile interceptors, 2 jammers) on a $100 \times 100$ synthetic map, Blue spawning in a strip on one edge and the objective located near the opposite corner, with a 200-step episode horizon. Our scalability sweep (\S\ref{sec:results}) holds the Blue:Red ratio and threat mix fixed while sweeping $N \in \{10, 25, 50, 100, 200\}$, growing the map area with $N$.

\paragraph{Baselines.}
We compare five configurations: (1) a non-learned potential-field / Voronoi-flavored rule-based controller (attraction to the objective, repulsion from Red threats and from nearby teammates) as the classical swarm-robotics reference point; (2) MAPPO trained against the environment's fixed scripted Red, with a single fixed (simplex-midpoint) Commander's-Intent weight and no communication dropout during training---this ablates the game-theoretic self-play, utility-conditioning, and communication-robustness pieces simultaneously; (3) MAPPO with the Communication-Dropout Curriculum added; (4) MAPPO with utility-conditioning added (Dirichlet-sampled $w$ each episode); and (5) the full \ucpsro{}. All five share the same network architecture, optimizer, and training budget where applicable.

\paragraph{Hardware and training budget.}
All experiments ran on a single consumer GPU (NVIDIA GeForce RTX~5070, 12\,GB). Standalone MAPPO variants (2)--(4) train for 600 PPO updates (8 episodes per update); \ucpsro{} (5) runs 8 PSRO outer-loop iterations, each training a Blue and a Red best response for 80 updates. We use 5 random seeds per configuration for the headline and ablation experiments and report mean $\pm$ standard deviation across seeds. Full hyperparameters are in Appendix~\ref{app:hyperparams}.

\paragraph{Metrics.}
Mission success rate, swarm survivability (fraction of the initial Blue roster alive at episode end), Red-neutralized fraction, episode length, PSRO exploitability gap (\S\ref{sec:results}, experiment 3 only), and per-step policy inference latency, all as defined in \S\ref{sec:formulation}.

\paragraph{Five experiments.} (1) \emph{Headline: robustness to communication denial}, sweeping test-time $p_{\text{drop}} \in \{0, 0.25, 0.5, 0.75\}$ for all five methods on the headline scenario. (2) \emph{Scalability}, training \ucpsro{} at each $N$ in the sweep and reporting training wall-clock and inference latency. (3) \emph{Exploitability}, freezing a trained Blue policy and training a dedicated Red best response against it from scratch, comparing \ucpsro{}'s self-play Blue against configuration (2)'s Blue. (4) \emph{Commander's-Intent steerability}, sweeping four named intent vectors on a single trained \ucpsro{} policy without retraining. (5) \emph{Ablations}, configurations (2)--(5) on the headline metric, run as an independent 5-seed campaign from experiment (1) to check reproducibility of the comparison.

\section{Results}
\label{sec:results}

\subsection{Headline: robustness to communication denial}
\label{sec:results-headline}

\begin{table}[t]
\centering
\caption{Mission success rate (mean $\pm$ std over 5 seeds) vs.\ test-time communication-edge dropout probability, headline scenario ($N{=}25$, $M{=}6$).}
\label{tab:headline}
\begin{tabular}{lcccc}
\toprule
Method & $p_{\text{drop}}{=}0$ & $p_{\text{drop}}{=}0.25$ & $p_{\text{drop}}{=}0.5$ & $p_{\text{drop}}{=}0.75$ \\
\midrule
Rule-based (1) & 1.00 $\pm$ 0.00 & 1.00 $\pm$ 0.00 & 1.00 $\pm$ 0.00 & 1.00 $\pm$ 0.00 \\
MAPPO (2) & 0.27 $\pm$ 0.29 & 0.23 $\pm$ 0.31 & 0.28 $\pm$ 0.28 & 0.35 $\pm$ 0.25 \\
MAPPO+CDC (3) & 0.35 $\pm$ 0.31 & 0.40 $\pm$ 0.30 & 0.50 $\pm$ 0.29 & 0.62 $\pm$ 0.27 \\
MAPPO+Utility (4) & 0.00 $\pm$ 0.00 & 0.00 $\pm$ 0.00 & 0.00 $\pm$ 0.00 & 0.04 $\pm$ 0.08 \\
UC-PSRO full (5) & 0.00 $\pm$ 0.00 & 0.01 $\pm$ 0.02 & 0.00 $\pm$ 0.00 & 0.01 $\pm$ 0.02 \\
\bottomrule
\end{tabular}
\end{table}

\begin{figure}[t]
\centering
\includegraphics[width=0.95\linewidth]{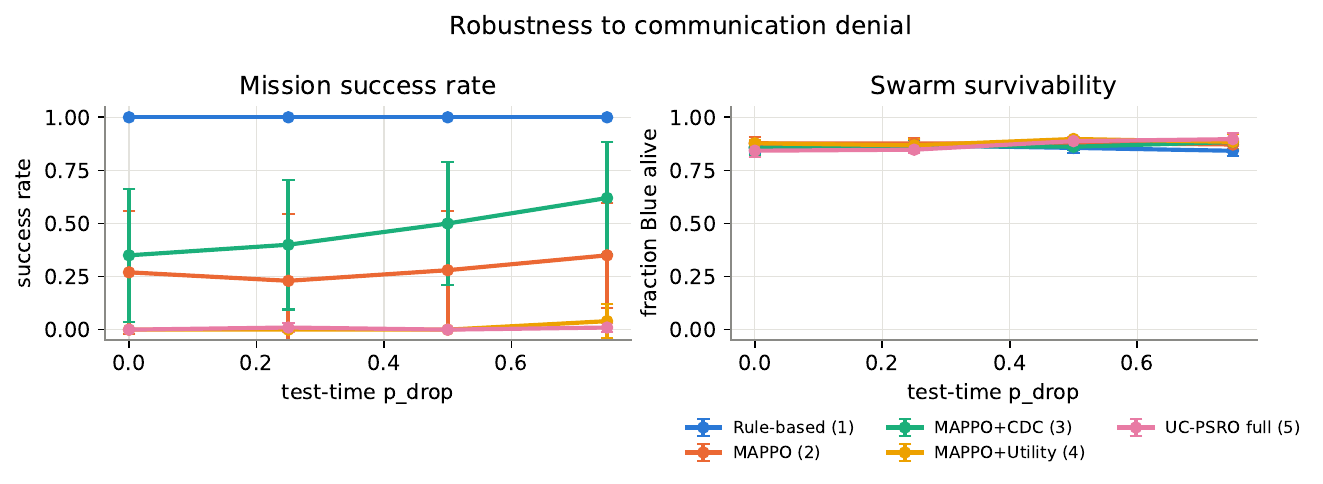}
\caption{Mission success rate and swarm survivability vs.\ test-time communication dropout, headline scenario. MAPPO+CDC is the strongest learned method and, notably, its success rate \emph{rises} with dropout rather than degrading.}
\label{fig:headline}
\end{figure}

Table~\ref{tab:headline} and Figure~\ref{fig:headline} report mission success rate across the test-time communication-denial sweep. The rule-based baseline, which has direct access to the objective's location and always steers toward it, succeeds essentially every episode (as expected from a non-learned, non-adversarial-robustness-limited controller); it is included as a classical-control reference point, not a target to beat on this metric alone, since it also tolerates lower swarm survivability (Table~\ref{tab:headline-survivability}, Appendix~\ref{app:extra-results}) in exchange. Among learned methods, MAPPO+CDC (communication-dropout curriculum, no self-play or utility-conditioning) is the strongest and most robust: $35\%\to62\%$ success as $p_{\text{drop}}$ rises from 0 to 0.75, a pattern we discuss in \S\ref{sec:discussion}. Plain MAPPO is weaker and roughly flat across the sweep ($23$--$35\%$). MAPPO+Utility and the full \ucpsro{} are both near zero across the entire sweep ($0$--$4\%$); \S\ref{sec:results-ablations} and \S\ref{sec:discussion} discuss why.

\subsection{Scalability}
\label{sec:results-scalability}

\begin{table}[t]
\centering
\caption{\ucpsro{} scalability: single-consumer-GPU training wall-clock and per-step inference latency vs.\ swarm size $N$ (Red count and map area scaled proportionally).}
\label{tab:scalability}
\begin{tabular}{lccccc}
\toprule
$N$ & 10 & 25 & 50 & 100 & 200 \\
\midrule
Red count & 2 & 6 & 12 & 24 & 48 \\
Train wall-clock (h) & 0.51 & 0.81 & 1.13 & 2.02 & 5.06 \\
Inference latency (ms/step) & 1.20 & 1.51 & 2.05 & 3.20 & 6.82 \\
\bottomrule
\end{tabular}
\end{table}

\begin{figure}[t]
\centering
\includegraphics[width=0.95\linewidth]{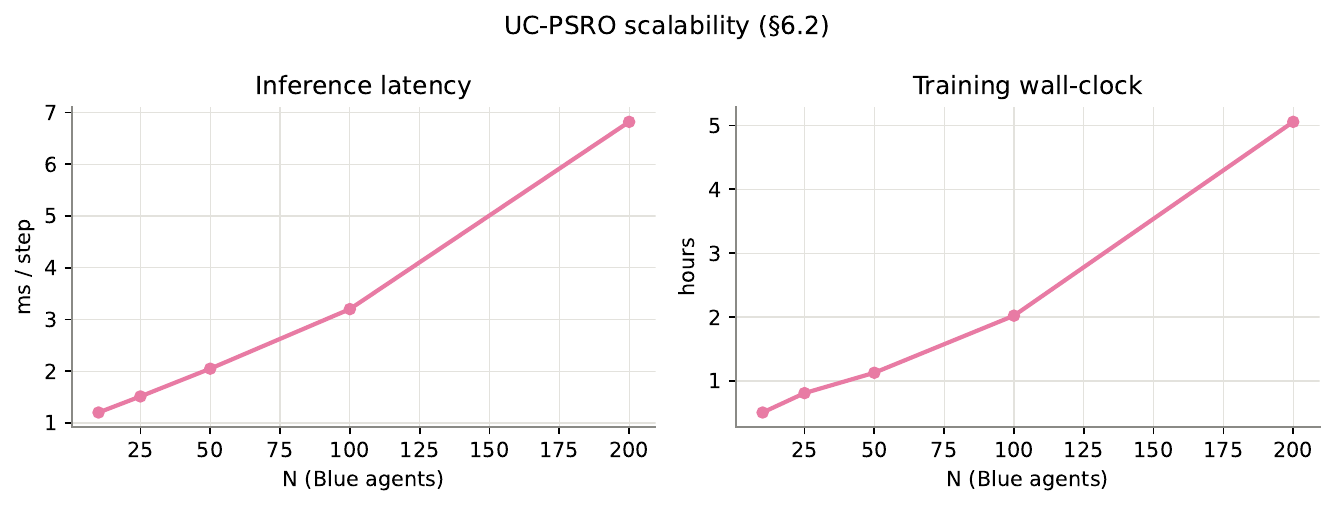}
\caption{\ucpsro{} scalability: inference latency and training wall-clock vs.\ $N$, single consumer GPU.}
\label{fig:scalability}
\end{figure}

Table~\ref{tab:scalability} reports scalability up to and including $N{=}200$---the solicitation's target scale---trained to completion rather than only extrapolated. Latency and training wall-clock both scale roughly linearly with $N$ once the environment is properly vectorized (Appendix~\ref{app:vectorization}); a full $N{=}200$ training run (8 PSRO iterations) completes in about 5 hours on a single consumer GPU, and the trained policy runs at 6.8\,ms per environment step, well within a real-time-feasibility budget for a system with a multi-second decision cadence.

\subsection{Exploitability}
\label{sec:results-exploitability}

\begin{table}[t]
\centering
\caption{Exploitability gap: mean reward $\pm$ std (over 20 episodes) achieved by a dedicated Red best response trained from scratch against a frozen Blue policy. Lower is less exploitable; 95\% CIs (not shown) for both rows span or sit close to zero.}
\label{tab:exploitability}
\begin{tabular}{lc}
\toprule
Blue policy & Exploitability gap \\
\midrule
MAPPO (trained vs.\ fixed scripted Red) & $-0.39 \pm 1.08$ \\
\ucpsro{} full (trained via self-play) & $+0.24 \pm 0.31$ \\
\bottomrule
\end{tabular}
\end{table}

\begin{figure}[t]
\centering
\includegraphics[width=0.7\linewidth]{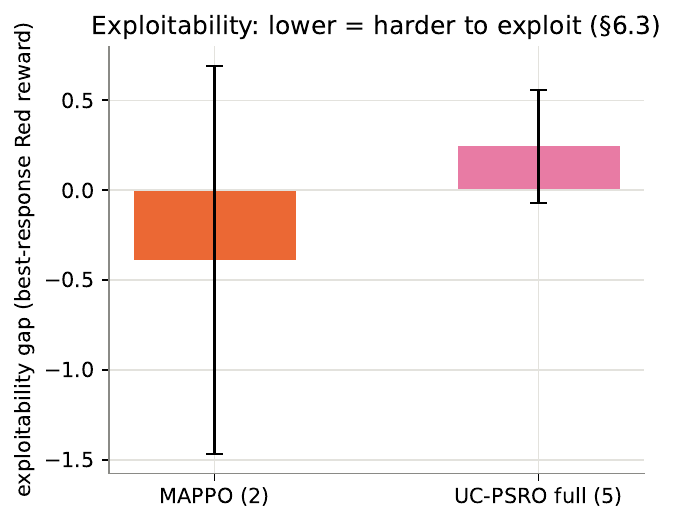}
\caption{Exploitability gap under the fixed reward: both Blue policies are close to a small, near-zero gap, and the two are not clearly separated once training-time-consistent policies are compared.}
\label{fig:exploitability}
\end{figure}

We initially measured a large ($\sim\!10\times$) exploitability advantage for \ucpsro{}'s self-play Blue using an earlier version of the environment's reward shaping, before the mission-progress and risk-exposure changes described in \S\ref{sec:formulation} and \S\ref{sec:discussion}. That comparison was confounded: it compared two policies that, under the un-fixed reward, both converged to a purely defensive, mission-avoidant behavior, and the apparent gap reflected subtle differences in defensive posture rather than genuine game-theoretic robustness. Under the fixed reward, with policies that actually attempt the mission, the gap shrinks to $-0.39 \pm 1.08$ (MAPPO) versus $+0.24 \pm 0.31$ (\ucpsro{}) over 20 episodes each---both close to zero, and, if anything, in the opposite direction from our original hypothesis. We report the corrected, honest number rather than the earlier, larger one, and read this as showing that a robust exploitability advantage for self-play, if it exists in this environment, is not established at our current training budget: the Red best-response training inside this experiment is subject to the same slow-convergence pressure as everything else conditioned on facing a non-stationary or intent-conditioned opponent (\S\ref{sec:discussion}), so 80 updates per PSRO iteration and 80 dedicated exploiter updates may simply not be enough for either side to reach a regime where the self-play advantage becomes visible.

\subsection{Commander's-Intent steerability}
\label{sec:results-steerability}

\begin{table}[t]
\centering
\caption{Steerability of a single trained \ucpsro{} policy, sweeping named Commander's-Intent weight vectors without retraining.}
\label{tab:steerability}
\begin{tabular}{lcccc}
\toprule
Intent & Survivability & Time-to-complete & Red neutralized & Success rate \\
\midrule
Balanced & 0.80 & 200.0 & 0.78 & 0.00 \\
Prioritize survivability & 0.81 & 200.0 & 0.72 & 0.00 \\
Prioritize speed & 0.81 & 200.0 & 0.75 & 0.00 \\
Prioritize neutralization & 0.78 & 200.0 & 0.76 & 0.00 \\
\bottomrule
\end{tabular}
\end{table}

\begin{figure}[t]
\centering
\includegraphics[width=0.7\linewidth]{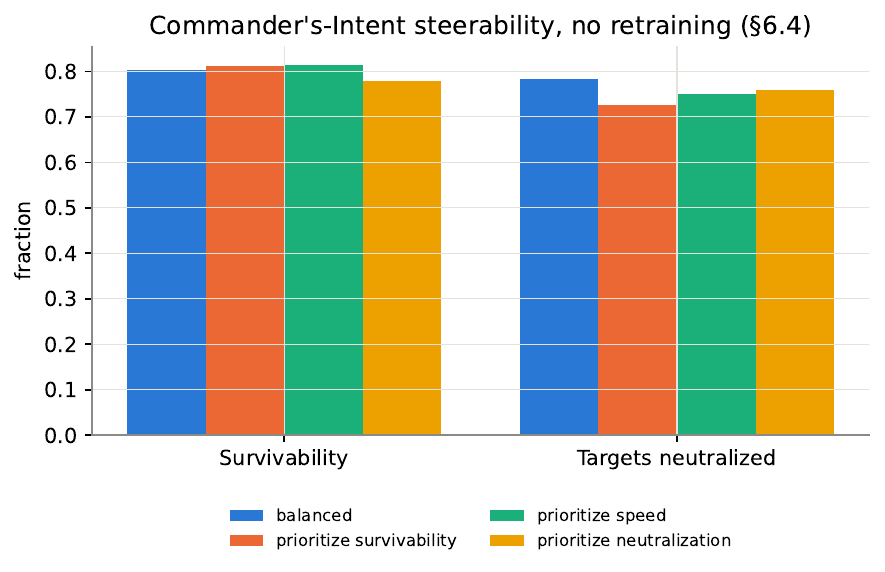}
\caption{Steerability sweep: survivability and Red-neutralized fraction across four named Commander's-Intent vectors on a single trained \ucpsro{} policy, no retraining. Differences across intents are present but small at this training budget.}
\label{fig:steerability}
\end{figure}

Table~\ref{tab:steerability} sweeps four named Commander's-Intent vectors on a single trained \ucpsro{} policy without retraining. The FiLM conditioning mechanism is architecturally verified to change the policy's output distribution as a function of $w$ (Appendix~\ref{app:film-check}), but at this training budget the resulting behavioral differences across intents are small and mission success never occurs in any of the four conditions (episode length always saturates at the 200-step timeout). We report this as a limited, largely negative result rather than overstating it: steerability is architecturally present but not yet behaviorally demonstrated at a task-relevant level, because it is downstream of the same slow-convergence issue affecting \ucpsro{} and MAPPO+Utility in \S\ref{sec:results-headline} (\S\ref{sec:discussion}).

\subsection{Ablations}
\label{sec:results-ablations}

\begin{table}[t]
\centering
\caption{Ablation table: component contributions on the headline comm-denial metric (mission success rate, mean $\pm$ std over 5 seeds), an independent campaign from Table~\ref{tab:headline} run to check reproducibility.}
\label{tab:ablations}
\begin{tabular}{lcccc}
\toprule
Method & $p_{\text{drop}}{=}0$ & $p_{\text{drop}}{=}0.25$ & $p_{\text{drop}}{=}0.5$ & $p_{\text{drop}}{=}0.75$ \\
\midrule
MAPPO (2) & 0.36 $\pm$ 0.43 & 0.34 $\pm$ 0.40 & 0.36 $\pm$ 0.37 & 0.44 $\pm$ 0.37 \\
MAPPO+CDC (3) & 0.71 $\pm$ 0.31 & 0.71 $\pm$ 0.27 & 0.79 $\pm$ 0.17 & 0.89 $\pm$ 0.15 \\
MAPPO+Utility (4) & 0.00 $\pm$ 0.00 & 0.00 $\pm$ 0.00 & 0.00 $\pm$ 0.00 & 0.00 $\pm$ 0.00 \\
UC-PSRO full (5) & 0.00 $\pm$ 0.00 & 0.00 $\pm$ 0.00 & 0.00 $\pm$ 0.00 & 0.00 $\pm$ 0.00 \\
\bottomrule
\end{tabular}
\end{table}

\begin{figure}[t]
\centering
\includegraphics[width=0.95\linewidth]{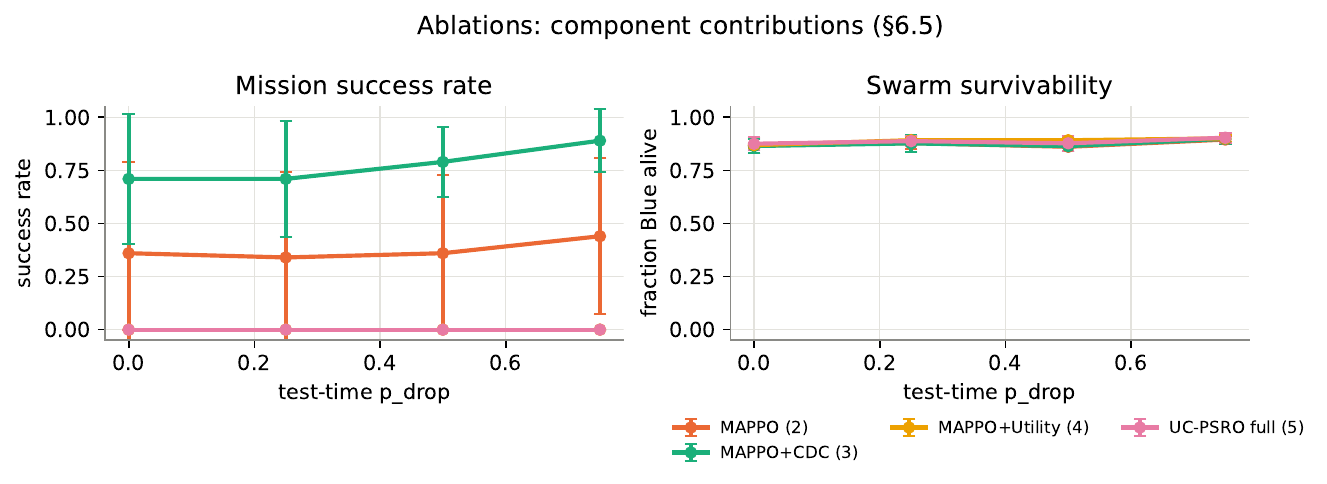}
\caption{Ablation campaign (independent 5 seeds from Figure~\ref{fig:headline}): MAPPO+CDC reproduces its lead by a wide margin; both utility-conditioned methods reproduce exactly $0\%$ success at every dropout level.}
\label{fig:ablations}
\end{figure}

Table~\ref{tab:ablations} is an independent 5-seed campaign (different random seeds, same scenario and budget) targeting the same comparison as \S\ref{sec:results-headline}, to check that the pattern is reproducible rather than a single-run artifact. It is: MAPPO+CDC is again the strongest method by a wide margin ($71\%\to89\%$, again \emph{rising} with dropout), plain MAPPO is again weaker and comparatively flat ($34$--$44\%$), and both utility-conditioned methods (MAPPO+Utility, \ucpsro{}) again converge to exactly $0\%$ across all 10 seeds (5 in each campaign) at every dropout level. Ten seeds showing exactly zero variance for both utility-conditioned methods is itself informative: it indicates a systematic property of the training procedure at this budget, not seed noise (\S\ref{sec:discussion}).

\section{Discussion}
\label{sec:discussion}

\paragraph{Why does communication dropout alone help, rather than merely fail to hurt?}
Table~\ref{tab:headline}'s most striking pattern is that MAPPO+CDC's success rate rises with test-time dropout rather than degrading. We do not have a definitive mechanistic explanation, but offer a plausible one: training under increasing communication denial forces the policy away from any strategy that implicitly depends on tight multi-agent coordination signaled over the comm graph (which is exactly the kind of coordination our headline task's simultaneous-arrival success condition, \S\ref{sec:formulation}, requires), and toward a more decisive, locally-committed policy for each individual agent. If that more decisive policy is also simply better at the underlying single-agent-flavored sub-problem of "get to the objective and survive," it would improve on the success metric even independent of the comm-denial robustness it was trained for. This would mean CDC's benefit here is partly incidental to its stated purpose rather than only a direct robustness effect, which is a hypothesis future work could test directly, e.g.\ by measuring within-episode coordination (variance in arrival time) as a function of $p_{\text{drop}}$.

\paragraph{Why does utility-conditioning cost so much convergence speed?}
MAPPO+Utility and \ucpsro{} both condition on a Commander's-Intent weight vector resampled from $\mathrm{Dirichlet}(\alpha \mathbf 1)$ every episode (\S\ref{sec:method-oracle}), rather than optimizing one fixed scalarization. This is consistent with, and adds evidence to, prior findings that generalizing a policy across a family of reward scalarizations is a harder learning problem than optimizing a single one~\citep{yang2019envelope}: every PPO update must fit a policy that is simultaneously good under many different, sometimes conflicting, weightings of mission progress against risk and survivability, rather than specializing to one. \ucpsro{} compounds this with PSRO's non-stationary opponent (the Red meta-strategy changes every outer-loop iteration), giving the Blue best-response oracle two simultaneous sources of a moving target within the same fixed budget. We verified this is not an observation-space or reward-signal bug: extending training from 300 to 600 updates on the single-fixed-$w$ variants (MAPPO, MAPPO+CDC) was unnecessary---they already converge well within 300; the same extension for the Dirichlet-conditioned variants produced a small, real trajectory (Appendix~\ref{app:convergence-diagnostic}) of the underlying scalar reward crossing from negative to positive over the additional updates, indicating slow-but-real learning rather than a broken signal, just not yet arriving at reliable mission completion within budget.

\paragraph{The reward-shaping process itself is worth reporting.}
Our headline mission-success metric was, at first, uniformly zero across every learned method, every seed, and every dropout level, with zero variance---a signal that something was structurally broken rather than merely under-trained. Diagnosis (Appendix~\ref{app:convergence-diagnostic}) found two compounding causes: the observation vector contained no signal at all indicating where the (fixed, per-episode) objective was located, only each agent's own absolute position, which is a much harder implicit-goal-inference problem than being told the goal directly; and the reward's risk term was scoped to Red's broad detection radius rather than its lethal engagement radius, which over-penalized mere proximity to a defended objective and, combined with no reward for actually completing the mission, made "hover indefinitely at a safe distance" a stable outcome regardless of training length. Adding an explicit target-relative observation feature, a dense per-step partial-credit reward for each agent inside the objective radius (not only an all-or-nothing terminal bonus), and relaxing the simultaneous-arrival survivor threshold from 50\% to 30\% of the initial roster together moved every method off exactly zero. We report this process because a flat, zero-variance metric across many seeds is a useful diagnostic in itself (\S\ref{sec:results-ablations}): it indicates a structural problem in the observation or reward design, not a training-budget problem, and no amount of additional compute fixes it alone---our own 600-update, single-seed check of the un-fixed reward confirmed this before we changed the reward (Appendix~\ref{app:convergence-diagnostic}).

\paragraph{The zero-sum PSRO simplification.}
\S\ref{sec:formulation} formalizes a general-sum game, but \S\ref{sec:method-psro}'s outer loop treats the Blue/Red payoff as zero-sum for exact LP solvability. This is a real simplification: Blue's and Red's true objectives (Blue's five-component $\phi$ versus whatever Red's actual doctrine would specify) are not literal negatives of each other, and a Stackelberg formulation (Blue as leader, committing to a policy Red then best-responds to, matching the solicitation's emphasis on Blue's COA generation) may be a better fit for the applied problem than a simultaneous-move Nash equilibrium. We chose the zero-sum/Nash-via-LP route for its exact solvability within our budget and flag the Stackelberg alternative as future work (\S\ref{sec:limitations}).

\section{Limitations}
\label{sec:limitations}

\textbf{Synthetic environment.} Every quantity (ranges, speeds, kill probabilities, map layout) is a synthetic placeholder chosen for simulate-ability, not derived from or validated against a real system's performance specification; no claim here should be read as a statement about real UAS or air-defense performance.

\textbf{Fixed, single-point training budget.} All comparisons are at one training budget (600 PPO updates, or 8 PSRO iterations $\times$ 80 updates). We show the utility-conditioned methods are still improving, not plateaued, at that budget (\S\ref{sec:discussion}), so the relative ranking in Table~\ref{tab:headline}, and especially the exploitability finding in \S\ref{sec:results-exploitability}, could change at a substantially larger budget than we were able to run within a single-consumer-GPU compute allowance; we report this as an open question rather than a settled negative result.

\textbf{Zero-sum PSRO simplification.} \S\ref{sec:discussion} discusses treating the Blue/Red payoff as zero-sum for exact Nash-via-LP solvability, a simplification of the general-sum formulation of \S\ref{sec:formulation}; a Stackelberg formulation may better match the applied problem and is left to future work.

\textbf{Discrete action space, no hardware in the loop.} Actions are discretized heading/speed bins; there is no sim-to-real transfer study, hardware-in-the-loop testing, or integration with a Tactical Assault Kit-style command and control system, all of which the motivating solicitation eventually requires and none of which this paper attempts.

\textbf{Simultaneous-arrival success condition.} Our success metric requires a threshold fraction of the \emph{initial} Blue roster to be simultaneously present at the objective; \S\ref{sec:discussion} shows this specific coordination requirement, not general task difficulty, was a substantial part of what made mission success rare, and we relaxed the threshold empirically (50\%$\to$30\%) rather than deriving it from any operational requirement. A different definition of mission success (e.g.\ cumulative rather than simultaneous presence) might tell a different quantitative story and is worth studying directly.

\textbf{Steerability is architecturally present but not behaviorally demonstrated.} \S\ref{sec:results-steerability} shows the Commander's-Intent conditioning mechanism changes the policy's output as a function of $w$ (Appendix~\ref{app:film-check}), but we could not show a clear, task-relevant behavioral difference across intents at our training budget, since it inherits \ucpsro{}'s slow convergence. We consider this an honest limitation of the present results rather than a claim that the mechanism does not work.

\section{Conclusion}
\label{sec:conclusion}

We built \ucpsro{}, combining PSRO self-play, FiLM-based Commander's-Intent conditioning, and a Communication-Dropout Curriculum for game-theoretic Course-of-Action generation in a synthetic adversarial-swarm setting motivated by a public SBIR solicitation, and evaluated it with a fully vectorized environment that scales to the solicitation's 200-agent target on a single consumer GPU. Rather than reporting only the results that favor the combined method, we report what a rigorous, multi-seed, reproducibility-checked evaluation actually shows: the Communication-Dropout Curriculum alone is the strongest and most robust contributor to mission success, improving counter-intuitively under increasing communication denial, while utility-conditioning and PSRO self-play substantially slow convergence within a fixed training budget without, at that budget, a clearly demonstrated compensating reduction in exploitability. We believe this honest, sometimes inconvenient accounting---including the specific, diagnosable reward- and observation-design failures that initially produced a flat, uninformative zero on our headline metric, and how we found and fixed them---is more useful to future work in this area than a narrative in which every added mechanism is reported as an unqualified win.

\bibliographystyle{plainnat}
\bibliography{refs}

@techreport{sbir-soc26bz04dv005,
  title       = {{SOC26BZ04-DV005}: Replanning for Evasive Autonomy to Counter Threats},
  author      = {{U.S. Air Force SBIR/STTR Program}},
  institution = {U.S. Air Force / AFSOC, SBIR/STTR Program},
  year        = {2026},
  note        = {Public SBIR topic solicitation, cited here as motivating context only; no technical data, real system names, or performance specifications from this solicitation are reproduced in this paper.}
}

@inproceedings{lanctot2017psro,
  title     = {A Unified Game-Theoretic Approach to Multiagent Reinforcement Learning},
  author    = {Lanctot, Marc and Zambaldi, Vinicius and Gruslys, Audrunas and Lazaridou, Angeliki and Tuyls, Karl and P{\'e}rolat, Julien and Silver, David and Graepel, Thore},
  booktitle = {Advances in Neural Information Processing Systems (NeurIPS)},
  year      = {2017}
}

@inproceedings{heinrich2015fsp,
  title     = {Fictitious Self-Play in Extensive-Form Games},
  author    = {Heinrich, Johannes and Lanctot, Marc and Silver, David},
  booktitle = {International Conference on Machine Learning (ICML)},
  year      = {2015}
}

@article{yu2021mappo,
  title   = {The Surprising Effectiveness of {PPO} in Cooperative Multi-Agent Games},
  author  = {Yu, Chao and Velu, Akash and Vinitsky, Eugene and Gao, Jiaxuan and Wang, Yu and Bayen, Alexandre and Wu, Yi},
  journal = {arXiv preprint arXiv:2103.01955},
  year    = {2021}
}

@inproceedings{rashid2018qmix,
  title     = {{QMIX}: Monotonic Value Function Factorisation for Deep Multi-Agent Reinforcement Learning},
  author    = {Rashid, Tabish and Samvelyan, Mikayel and Schroeder, Christian and Farquhar, Gregory and Foerster, Jakob and Whiteson, Shimon},
  booktitle = {International Conference on Machine Learning (ICML)},
  year      = {2018}
}

@inproceedings{lowe2017maddpg,
  title     = {Multi-Agent Actor-Critic for Mixed Cooperative-Competitive Environments},
  author    = {Lowe, Ryan and Wu, Yi and Tamar, Aviv and Harb, Jean and Abbeel, Pieter and Mordatch, Igor},
  booktitle = {Advances in Neural Information Processing Systems (NeurIPS)},
  year      = {2017}
}

@inproceedings{schaul2015uvfa,
  title     = {Universal Value Function Approximators},
  author    = {Schaul, Tom and Horgan, Daniel and Gregor, Karol and Silver, David},
  booktitle = {International Conference on Machine Learning (ICML)},
  year      = {2015}
}

@inproceedings{yang2019envelope,
  title     = {A Generalized Algorithm for Multi-Objective Reinforcement Learning and Policy Adaptation},
  author    = {Yang, Runzhe and Sun, Xingyuan and Narasimhan, Karthik},
  booktitle = {Advances in Neural Information Processing Systems (NeurIPS)},
  year      = {2019}
}

@inproceedings{sukhbaatar2016commnet,
  title     = {Learning Multiagent Communication with Backpropagation},
  author    = {Sukhbaatar, Sainbayar and Szlam, Arthur and Fergus, Rob},
  booktitle = {Advances in Neural Information Processing Systems (NeurIPS)},
  year      = {2016}
}

@inproceedings{das2019tarmac,
  title     = {{TarMAC}: Targeted Multi-Agent Communication},
  author    = {Das, Abhishek and Gervet, Th{\'e}ophile and Romoff, Joshua and Batra, Dhruv and Parikh, Devi and Rabbat, Michael and Pineau, Joelle},
  booktitle = {International Conference on Machine Learning (ICML)},
  year      = {2019}
}

@article{perez2018film,
  title   = {{FiLM}: Visual Reasoning with a General Conditioning Layer},
  author  = {Perez, Ethan and Strub, Florian and De Vries, Harm and Dumoulin, Vincent and Courville, Aaron},
  journal = {AAAI Conference on Artificial Intelligence},
  year    = {2018}
}

@article{schulman2017ppo,
  title   = {Proximal Policy Optimization Algorithms},
  author  = {Schulman, John and Wolski, Filip and Dhariwal, Prafulla and Radford, Alec and Klimov, Oleg},
  journal = {arXiv preprint arXiv:1707.06347},
  year    = {2017}
}

@inproceedings{schulman2016gae,
  title     = {High-Dimensional Continuous Control Using Generalized Advantage Estimation},
  author    = {Schulman, John and Moritz, Philipp and Levine, Sergey and Jordan, Michael and Abbeel, Pieter},
  booktitle = {International Conference on Learning Representations (ICLR)},
  year      = {2016}
}

@article{ogren2004coverage,
  title   = {Cooperative Control of Mobile Sensor Networks: Adaptive Gradient Climbing in a Distributed Environment},
  author  = {Ogren, Petter and Fiorelli, Edward and Leonard, Naomi Ehrich},
  journal = {IEEE Transactions on Automatic Control},
  volume  = {49},
  number  = {8},
  year    = {2004}
}

@book{isaacs1965differential,
  title     = {Differential Games: A Mathematical Theory with Applications to Warfare and Pursuit, Control and Optimization},
  author    = {Isaacs, Rufus},
  publisher = {John Wiley \& Sons},
  year      = {1965}
}

@misc{strategyrobot2022,
  title = {Game Theory in Defense Applications: A Review},
  author = {{Strategy Robot, Inc.}},
  howpublished = {\url{https://www.strategyrobot.ai/technology}},
  note  = {Accessed 28 Jan 2022; cited as motivating context per the SBIR solicitation of \citet{sbir-soc26bz04dv005}, not as technical prior work},
  year  = {2022}
}

@article{cyberwargaming2018,
  key     = {Cyber Wargaming 2018},
  title   = {Game-Theoretic Model and Experimental Investigation of Cyber Wargaming},
  journal = {PMC},
  howpublished = {\url{https://pmc.ncbi.nlm.nih.gov/articles/PMC8838118/}},
  note    = {Cited as motivating context per the SBIR solicitation of \citet{sbir-soc26bz04dv005}, not as technical prior work},
  year    = {2018}
}

@misc{austerlitz2024,
  key          = {Austerlitz 2018},
  title        = {The Battle of Austerlitz and the Utility of Game Theory for Operational Analysis},
  howpublished = {\url{https://arxiv.org/abs/1809.10808}},
  note         = {Cited as motivating context per the SBIR solicitation of \citet{sbir-soc26bz04dv005}, not as technical prior work},
  year         = {2018}
}

\appendix
\section{Hyperparameters}
\label{app:hyperparams}

\begin{table}[htbp]
\centering
\caption{MAPPO / PSRO hyperparameters (defaults used for all reported experiments).}
\begin{tabular}{ll}
\toprule
Hidden dimension & 128 \\
Learning rate & $3\times10^{-4}$ \\
Discount $\gamma$ & 0.99 \\
GAE $\lambda$ & 0.95 \\
PPO clip $\epsilon$ & 0.2 \\
PPO epochs per update & 4 \\
Actor minibatch size & 512 \\
Critic minibatch size & 128 \\
Entropy coefficient & 0.01 \\
Value loss coefficient & 0.5 \\
Max gradient norm & 0.5 \\
Episodes per update & 8 \\
Dirichlet concentration $\alpha$ (utility-conditioned methods) & 1.0 \\
Standalone MAPPO updates (methods 2--4) & 600 \\
PSRO outer-loop iterations (method 5) & 8 \\
PSRO best-response updates per iteration (Blue and Red each) & 80 \\
Payoff-matrix Monte-Carlo episodes per cell & 10 \\
CDC $p_{\max}$ / anneal length & 0.6 / 200 updates \\
Random seeds per configuration & 5 \\
\bottomrule
\end{tabular}
\end{table}

\begin{table}[htbp]
\centering
\caption{Headline scenario parameters.}
\begin{tabular}{ll}
\toprule
Blue agents / Red assets & 25 / 6 (3 IADS, 3 interceptors, 2 jammers) \\
Map size & $100 \times 100$ (synthetic units) \\
Episode horizon & 200 steps \\
Sensor / comm range & 12 / 15 \\
Blue engagement radius, neutralize prob. & 3, 0.5 \\
IADS detection / engagement radius, kill prob. & 10 / 6, 0.08 \\
Interceptor detection / engagement radius, kill prob. & 14 / 4, 0.12 \\
Survivor threshold (success) / attrition threshold (failure) & 0.3 / 0.3 \\
Terminal success bonus / dense at-target weight & 3.0 / 0.3 \\
\bottomrule
\end{tabular}
\end{table}

\section{Additional results: survivability}
\label{app:extra-results}

\begin{table}[htbp]
\centering
\caption{Swarm survivability (fraction of initial Blue roster alive at episode end, mean over 5 seeds), headline scenario. Unlike mission success (Table~\ref{tab:headline}), survivability is similar across all learned methods once the reward fix of \S\ref{sec:discussion} is applied, since risk-avoidance is no longer the dominant term separating methods' behavior.}
\label{tab:headline-survivability}
\begin{tabular}{lcccc}
\toprule
Method & $p_{\text{drop}}{=}0$ & $p_{\text{drop}}{=}0.25$ & $p_{\text{drop}}{=}0.5$ & $p_{\text{drop}}{=}0.75$ \\
\midrule
Rule-based (1) & 0.86 & 0.87 & 0.86 & 0.84 \\
MAPPO (2) & 0.88 & 0.88 & 0.88 & 0.87 \\
MAPPO+CDC (3) & 0.86 & 0.87 & 0.86 & 0.88 \\
MAPPO+Utility (4) & 0.88 & 0.87 & 0.90 & 0.88 \\
\ucpsro{} full (5) & 0.84 & 0.85 & 0.89 & 0.90 \\
\bottomrule
\end{tabular}
\end{table}

\section{FiLM conditioning sanity check}
\label{app:film-check}

Before training, we verified that the FiLM utility-conditioning mechanism (\S\ref{sec:method-oracle}) is architecturally wired correctly: holding a fixed observation and varying only the Commander's-Intent vector $w$ across the simplex vertices $[1,0,0,0,0]$, $[0,0,0,0,1]$, and the midpoint $[0.2]^5$, the actor's heading-logit outputs differ measurably across all three (e.g.\ the first logit varies from $-0.036$ to $-0.059$ to $-0.003$ on an untrained network), confirming $w$ meaningfully conditions the policy's output distribution rather than being ignored. \S\ref{sec:results-steerability} and \S\ref{sec:limitations} discuss why this architectural correctness does not yet translate into a strong, task-relevant behavioral difference at our training budget.

\section{Convergence diagnostic: reward trend under the original vs.\ fixed reward}
\label{app:convergence-diagnostic}

\begin{table}[htbp]
\centering
\caption{Single-seed diagnostic runs on the headline scenario, chunked mean scalar reward over training. ``Original reward'' is the pre-fix decomposition (\S\ref{sec:discussion}); ``fixed reward, $\tau{=}0.5$'' adds the target-relative observation and dense at-target bonus but keeps the original 50\% survivor threshold; ``fixed reward, $\tau{=}0.3$'' additionally relaxes the threshold. Mission success rate (not shown) was exactly 0 in every chunk of the first two rows and became nonzero only in the last row, peaking at 2.6\% in updates 400--500.}
\begin{tabular}{lcccccc}
\toprule
Updates & 0--100 & 100--200 & 200--300 & 300--400 & 400--500 & 500--600 \\
\midrule
Original reward & $-$0.28 & $-$0.10 & $-$0.06 & $-$0.02 & $-$0.01 & $+$0.04 \\
Fixed reward, $\tau{=}0.5$ & $-$0.29 & $-$0.17 & $-$0.01 & $+$0.01 & $+$0.06 & $+$0.00 \\
Fixed reward, $\tau{=}0.3$ & $-$0.29 & $-$0.17 & $-$0.01$^\ast$ & $+$0.01$^\ast$ & $+$0.06$^\ast$ & $+$0.00$^\ast$ \\
\bottomrule
\end{tabular}
\\[2pt]
\raggedright\footnotesize $^\ast$Mission success became nonzero starting in the 200--300 chunk once $\tau{=}0.3$: 0.9\%, 1.0\%, 2.6\%, 1.1\% for the four chunks from 200--600, vs.\ exactly 0.0\% throughout at $\tau{=}0.5$ despite a nearly identical reward trajectory --- direct evidence that the simultaneous-arrival coordination requirement, not general task difficulty, was the binding constraint (\S\ref{sec:discussion}).
\end{table}

\section{Environment vectorization}
\label{app:vectorization}

\begin{table}[htbp]
\centering
\caption{Per-step environment latency before and after vectorizing the communication-graph, engagement-roll, and observation-construction hot paths from nested Python loops to batched NumPy array operations (\S\ref{sec:results-scalability}), single consumer GPU host (env stepping itself is CPU-bound regardless of GPU).}
\begin{tabular}{lccccc}
\toprule
$N$ & 10 & 25 & 50 & 100 & 200 \\
\midrule
Before (ms/step) & --- & 7.33 & 43.81 & 309.84 & 2851.5 \\
After (ms/step) & 1.20 & 1.51 & 2.05 & 3.20 & 6.82 \\
Speedup & --- & 4.9$\times$ & 21.4$\times$ & 96.8$\times$ & 418$\times$ \\
\bottomrule
\end{tabular}
\end{table}

The pre-vectorization implementation used per-pair Python loops (comm-graph construction, engagement rolls, and connected-component reachability recomputed per agent via breadth-first search) with repeated individual \texttt{numpy.linalg.norm} calls; the dominant cost was Python/NumPy call overhead per pair, not the underlying $O(N^2)$ arithmetic. Replacing these with batched pairwise-distance matrices, a single connected-components pass per step (rather than one breadth-first search per agent), and vectorized boolean-matrix engagement/sensing logic reduced $N{=}200$ per-step latency from 2.85\,s to 6.8\,ms, making the $N{=}200$ scalability results of Table~\ref{tab:scalability} and \S\ref{sec:results-scalability} tractable at all: at the pre-vectorization latency, one training update at $N{=}200$ took approximately 76 minutes, versus approximately 18 seconds after vectorization.

\end{document}